# Weightless Neural Network with Transfer Learning to Detect Distress in Asphalt

Suayder Milhomem[1], Tiago da Silva Almeida[2], Warley Gramacho da Silva[3], Edeilson Milhomem da Silva[4], Rafael Lima de Carvalho[5]

[1]Department of Computer Science, Federal University of Tocantins, BRAZIL
Emails: {[1]suayder, [2]tiagoalmeida, [3]wgramacho, [4]edeilson.milhomem, [5]rafael.lima}@uft.edu.br

**Abstract**— *The present paper shows a solution to the problem of automatic distress detection, more precisely the detection of holes in paved roads. To do so, the proposed solution uses a weightless neural network known as Wisard to decide whether an image of a road has any kind of cracks. In addition, the proposed architecture also shows how the use of transfer learning was able to improve the overall accuracy of the decision system. As a verification step of the research, an experiment was carried out using images from the streets at the Federal University of Tocantins, Brazil. The architecture of the developed solution presents a result of 85.71% accuracy in the dataset, proving to be superior to approaches of the state-of-the-art.*

## I. INTRODUCTION

In Brazil, most of the traffic is driven on asphalt roads. According to the August 2018 statical bulletin of the CNT (National Confederation of Transport), the freight transport matrix is 61% done by road traffic[1]. Due to such high traffic demand, the need for more asphalt roads is a reality in Brazil.

Despite the importance of road transport in Brazil, there is a serious problem due to the roads and highways condition. The detection of problems, such as cracks and other distress, is costly and usually is made manually by the transportation agencies. Unfortunately, the appearance of potholes in highways is faster than its detection. Consequently, the lack of better quality roads has led to consequences that outreach economic monetary values, such as fatal accidents which have claimed many lives that are irreplaceable.

The advances in the artificial intelligence techniques merged with image processing studies offers background for a new field, called computer vision. This area applies techniques capable of simulating the human brain's behavior through algorithms that try to emulate the visual perception of human beings. In this way, this paper shows the results of a system based on computer vision techniques to detect the existence of distress in images of a paved road. Therefore, the architecture of the presented solution uses a weightless neural network model called Wisard [1], along with transfer learning that is able to select features of input images.

The problem of distress detection has been investigated by many another researches. Ouma and Hahn [2] proposed the extraction and identification of incipient or micro-linear distresses in asphalt. To do so, their system is composed of three main approaches: Discrete Wavelet Transform to isolate and classify failures; Successive morphological Transformation Filtering to detect shapes in failures; Circular Radon Transform for angular-geometric orientation analysis for the identification and classification of distress types. As a result, their system achieved 83.2% of accuracy.

Rodopoulou [3] showed a method to detect patch in asphalt pavement through videos. The patches were detected through its visual features, which include closed contour and texture around them. The algorithm had 84% of precision in detection.

Gopalakrishnan et al. [4] use of Transfer Learning (TL) to classify whether there are or aren't damages on asphalt. For this task, the model used a VGG16 [5] that is a pre-trained neural network which is used to transfer learning task. For the classification task, it has been used a single layer perceptron. As result Gopalakrishnan obtained an precision of 0.9 on his dataset.

For automatic crack detection, Hoang, et al. [6] Applied two approaches, edge detection and a machine learning algorithm. For the edge detection, it has been used the Sobel and Canny algorithms, both using the Flower Pollination metaheuristics which is responsible for determining the threshold of the edges. The second algorithm uses a Convolutional Neural Network for detection and classification. The first approach, using the filters, both accuracies were 76.69% and 79.99% for the

---

[1]Source: http://www.transportes.gov.br/images/PAC_-_SITE_-_Fechado.pdf. Accessed on: December, 2018.

Canny and Sobel, respectively. The second approach resulted in an accuracy of 92.88%, when using the CNN. Moreover, Zhang [7] investigated a dataset composed of 800 low similarity images. Its classification system also used the transfer learning paradigm in order to extract generic knowledge from the first layer of a deep convolutional neural network. Such ANN was pre-trained using the ImageNet [8] database. After that, a fine-tuning strategy was used over the next ANN layers. Zhang's method reached values of recall as 0.951, precision as 0.847 and a F1-measure as 0.895.

The rest of this paper is structured as follows: The Section 2 presents the fundamentals of neural network and transfer learning. The proposed methodology is shown in Section 3. In addition, Section 4 summarizes the experiments and results of this proposed work. Lastly, in Section 5 there are some final remarks and future appointments of this research.

## II. ARTIFICIAL NEURAL NETWORKS AND TRANSFER LEARNING

Usually, regression and classification models are comprised as linear combinations of fixed basis functions. These models have analytical and computational properties that, in practical scenarios, are limited by dimensionality. This way, before the application of these models in large-scale problems, it is necessary to make the basis function adaptable to the data.

According to Bishop [9], for pattern recognition tasks, the most successful model to fit the data properly is the Artificial Neural Networks (ANN). This term began to be used in attempt of [10] to represent the perceptron mathematically which have encouraged the creation of a range of models of perceptrons and ANNs.

2.1 Deep Convolutional Neural Networks And Transfer Learning

A Convolutional Neural Network (CNN) is a type of ANN which is able to create models that are invariant to certain input transformations [11]. This ANN is probably the most well known Deep Learning (DL) model and the most used in computer vision tasks, particularly for image classification. The CNN combines three architectural mechanisms to guarantee the invariance of distortions. Specifically, these mechanisms are: local receptive fields, weight sharing and sub-sampling [12]. The basic building blocks of its architecture are convolutions, the pooling layers (sub-sampling) and the fully-connected layers (like *Multilayer Perceptron*). The basic structure of a CNN is shown in Fig 1.

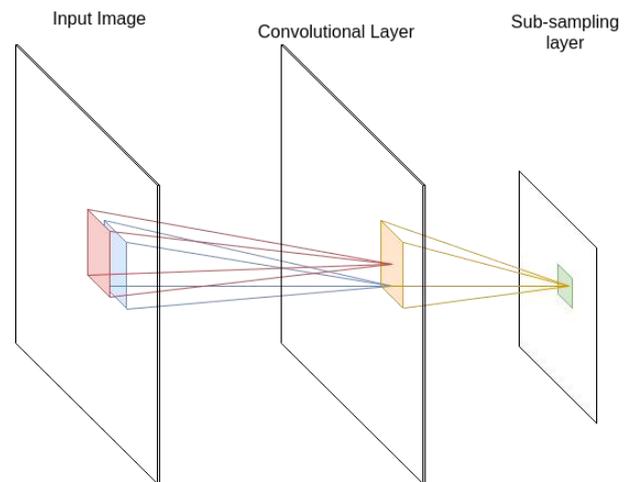

*Fig. 1: The general architecture of a CNN to classify images. Three different layers are normally used, being such: input layer to receive the image, convolutional layer and sub-sample layer.*

The convolutional layer is composed of units, which are organized into planes. Each such unit is called *feature map*. Each feature map is derived from small subregions of the input matrix. It is important to mention that the convolution operator in the image may have many parameters such as: stride, border mode and kernel size. The sub-sampling layer is responsible for making the network more invariant. In cases like that, the max-pooling operator is the most frequently used. Thus, the fully connected layers are at the end of the convolution layers and work in a similar way as the *Multilayer perceptron,* in order to learn weights to classify the data.

Deep Convolutional Neural Network (DCNN) usually requires large image dataset to achieve high values of accuracy. Unfortunately, in certain problems, there is no such amount of data. In such cases, the Transfer Learning (TL) methodology is an interesting alternative. Sinno [13] defines TL as: given a source domain $D_S$ and a learning task $T_S$, a target domain $D_T$ and a learning task $T_T$, transfer learning aims to help improve the learning of the target predictive function $f_T(\cdot)$ in $D_T$ using the knowledge in $D_S$ and $T_S$, where $D_S \neq D_T$, or $T_S \neq T_T$.

Thus, in practice when one wants to work with transfer learning, there are many off-the-shelf models. For example, the well-established neural networks which are pre-trained in large image datasets (such as *ImageNet*[2]). These models allow transfer their learning ability to a new classification scenario, instead of the need of training a DCNN from scratch. Some of these pre-trained models are found in frameworks like Tensorflow[3]. Only to illustrate, some examples include: VGG-16 [5], AlexNet

---

[2] http://www.image-net.org/
[3] https://www.tensorflow.org

[14], Inception V3 [15] and Xception [16]. This last one has been chosen in this research study.

The architecture of the Xception model [16] is composed of 36 convolutional layers forming its feature extraction basis. These layers are structured into 14 modules which are all interconnected, except for the first and last ones. In short, its architecture is a linear stack of depthwise separable convolution layers with residual connections.

2.2 The Wisard weightless neural network model

According to [17], Wisard is a pattern recognition machine based on neural principles (projected to be implemented in a hardware device). Wisard is composed of class units called discriminators. Each discriminator accommodates a set of RAM-based neurons. These neurons have a memory behaviour such as the Random Access Memory. The address is the binary pattern, and if there is information in such address, then the neuron fires. In summary, the main components of a Wisard device are: an address decoder, a group of memory registers, a register with the input data and a register for the output data.

The RAM-based neuron described earlier can be seen as a neuron in the sense that given an input pattern, it stores the desired output. In the training and possible re-training steps, the pattern of a given address can be overwritten by other values in subsequent training steps. Therefore, Wisard is different from an ordinary ANN because it does not need a sophisticated training algorithm.

Inside the Wisard, the simplest RAM-based network with generalization properties is called Discriminator. A Discriminator consists of a layer with $K$ RAMs, which deals with $N$ inputs. Each RAM stores $2^N$ words of one bit, and the single layer receives a binary pattern of KxN bits. The training consists in presenting the input patterns (binary) and storing the desired value (either 1 or 0) in memory locations addressed by the displayed pattern. Each RAM then stores part of the input pattern. When a test pattern is presented, the discriminator outputs the number of fired RAMs. Fig. 2 illustrates the architecture of a Wisard's discriminator.

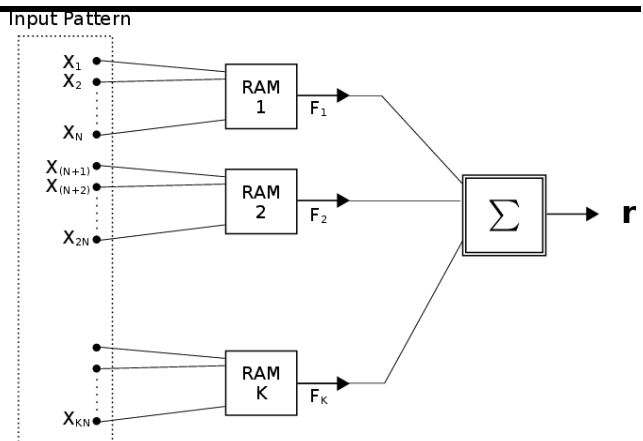

*Fig. 2: Discriminator Structure of Weightless Neural Network. The $X_{KN}$ means the input pattern for K RAMs, generating $F_K$ output function of each RAM [1].*

The Wisard consists of a multi-discriminator system where each of its discriminators are trained to recognize a different class.

### III. THE PROPOSED SOLUTION: WISARD WITH TRANSFER LEARNING

The proposed solution uses a pre-trained Xception deep convolutional neural network to pre-process the input image. When the input image is plugged in the Xception model, it produces an output vector with 2048 features. These features are then used as input to the weightless neural network Wisard, which calculates the sum of fired RAMs. In this case, if the sum of fired neurons is greater than a given threshold, the input image is classified as positive, i.e. it has any kind of distress.

In the proposed solution, the Xception has been pre-trained using general images from the *ImageNet* database. Furthermore, in order to apply the transfer learning, the last layer of the pre-trained Xception has been replaced by the Wisard model. The choice for the Wisard model as the final classifier is given by its high performance in both utilization as well as in learning [18]. Therefore, the overall system is illustrated in Fig. 3.

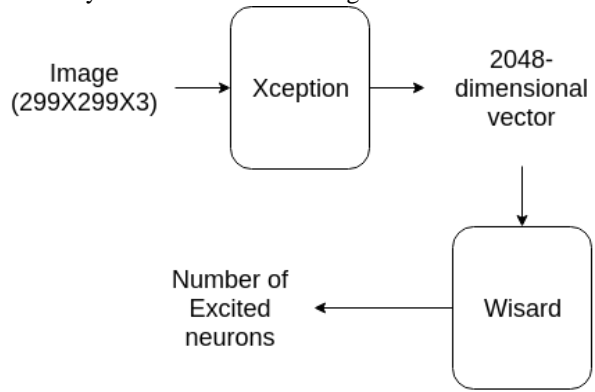

*Fig. 3: The architecture of the Wisard with Transfer Learning to the proposed solution. After the Xception, the images are dimensioned to be able to apply in the Wisard.*

One of the configuration parameters of the Wisard model is the amount of RAMs composing the network layer. As the last layer of the Xception network generates an output of 2048 features, the amount of RAMs in weightless networks depends only on the number of inputs defined for each RAM. In this case, the influence of this parameter was also investigated by this work. The evaluated values of the input size for each RAM is better addressed in Section 4.

Finally, it is also necessary to pre-process the input data used by the Wisard model. The standard input retina used by the Wisard is a binary data stream. In this way, the output of Xception has been binarized using the array average as threshold. The whole model was implemented using the Python programing language version 3.6 and using the Keras API for the deap neural networks.

## IV.    EXPERIMENTAL RESULTS AND DISCUSSION

In order to evaluate the proposed model, a dataset with real images has been collected. A drone model Phanton 4 was used to take the pictures. The drone flew over the streets of a University Campus. From this experiment, an amount of 78 (seventy-eight) images were collected, including 63 photos with potholes and 15 images without any problems. Each image was labelled manually. Fig. 4 shows some samples of the collected database.

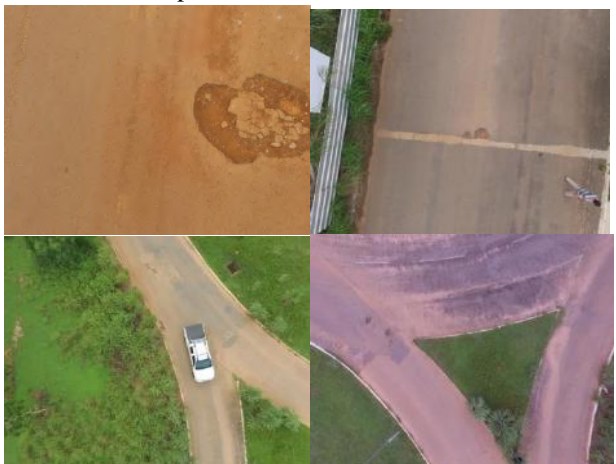

*Fig. 4: Samples of collected images of pavemented roads used in the tests of the proposed model.*

The test methodology is cross-validation, where the database was divided into 60 positive images (with potholes) for training and 3 positive images for test. Negative images (without potholes) were split in a way that the system trained with 14 samples and was evaluated using only 1 (one) negative sample for each subset, selected by the cross-validation procedure. Furthermore, the operational procedure for the test was Leave one out [18], a widely used methodology to evaluate classification systems. Therefore, each subset of tests is composed of 3 positive images and 1 negative one, while the remaining images are used for training. At the end of one cross-validation step, the test subset is inserted again into the whole set, while another subset of tests is separated to be evaluated by the system.

Using the aforementioned dataset, two experiments were carried out, which included: a) using the Wisard method without the transfer learning (labeled 'Without TL'); and b) using the architecture proposed in the section 3 (labeled 'With TL'). In addition, the related work proposed by Gopalakrishnan [4] was also implemented in order to compare with the reported findings of this work. Fig. 5 summarizes the results of the experiments a) and b).

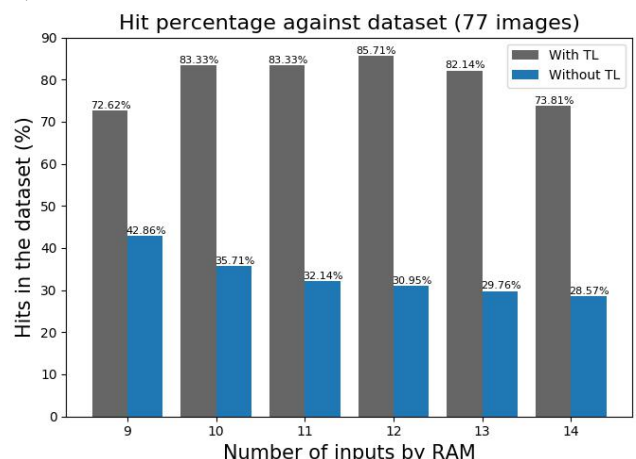

*Fig. 5: Accuracy values between Wisard ('Without TL) and the proposed system ('With TL'). In the horizontal axis is the input size for each RAM (a parameter for the Wisard neural model).*

As shown by the plot in Fig. 5, the use of transfer learning considerably improved the accuracy of the classification system. As noted in the Section 3, one of the Wisard parameters is the number of inputs used by each RAM-based neuron. Thus, Fig. 5 shows the sensitivity of the variation (9 to 14) of such parameter, in the whole classification performance. It could be seen that when this parameter is set to 12, the classification system reached 84.71% of accuracy, which was the best so far. On the other hand, when evaluating the system without transfer learning, the best result was 42.86%.

In order to verify the effectiveness of the model, a comparison was done with the related work proposed by Gopalakrishnan [4]. Its architecture consists of using the pre-trained neural network model VGG16 for transfer learning basis and as a classification model, a single-layer Perceptron neural network with 256 neurons in the input layer. Gopalakrishnan's model was trained and tested under the same test methodology as this work. The best results of each system are summarized in Tab. 1.

Tab. 1: Comparison of the best results on the test set images using the three evaluated systems.

| Classifier | Accuracy (%) |
|---|---|
| Wisard with TL | 85.71 |
| Wisard without TL | 42.86 |
| Gopalakrishnan Model | 74.51 |

As can be seen in Tab. 1 the proposed model was able to overperform one of the state-of-the art solutions, for the detection of potholes in paved roads. Since the database used by Gopalakrishnan [10] is not public, performance tests of the proposed work could not be evaluated.

## V. CONCLUSION

This research addressed the problem of classification of images with holes in paved roads using computer vision. In this work, the Wisard weightless neural network was evaluated as a detection system of images with and without holes. The proposed approach made use of a pre-trained convolutional neural network, known as Xception. With the knowledge of general images saved in the Xception, a detection system architecture that uses transfer learning to preprocess the input images and produce characteristics as the outcome. Such characteristics was then plugged in the input data of the Wisard network, which produced the label for the input image.

In order to evaluate the proposed architecture, an experiment was carried out, which included 77 images with and without potholes. In the conducted experiment, the transfer learning approach proved to be an effective solution, because it allowed to improve the general accuracy from 42.86% to 85.71%. In addition, when compared with a state of the art solution, the proposed approach overcame the previous one in 11.2%.

Some future insights include locating the potholes in the image, using a real-time detection solution.